%% file: main.tex
\documentclass[11pt]{article}
\usepackage{graphicx,amsmath,amsfonts,amssymb,bm,hyperref,url,breakurl,epsfig,epsf,color,fullpage,MnSymbol,mathbbol, fmtcount,semtrans
} 

\usepackage{titlesec}

\usepackage{tikz}
\usepackage{pgfplots}
\usepackage{stackengine}
\def\delequal{\mathrel{\ensurestackMath{\stackon[1pt]{=}{\scriptstyle\Delta}}}}

\usetikzlibrary{pgfplots.groupplots}

\usepackage{amsmath}
\usepackage{framed}
\usepackage{algorithm}
\usepackage[noend]{algpseudocode}

\titleformat{\paragraph}
{\normalfont\normalsize\bfseries}{\theparagraph}{1em}{}
\titlespacing*{\paragraph}
{0pt}{3.25ex plus 1ex minus .2ex}{1.5ex plus .2ex}

\usepackage{caption,subcaption}

\usepackage[bottom,hang,flushmargin]{footmisc} 

\usepackage{hyperref}
\definecolor{darkred}{RGB}{150,0,0}
\definecolor{darkgreen}{RGB}{0,150,0}
\definecolor{darkblue}{RGB}{0,0,200}
\hypersetup{colorlinks=true, linkcolor=darkred, citecolor=darkgreen, urlcolor=black}

\newtheorem{theorem}{Theorem}[section]
\newtheorem{lemma}[theorem]{Lemma}

\newtheorem{proposition}[theorem]{Proposition}

\newtheorem{remark}[subsection]{Remark}

\newcommand{\R}{\mathbb{R}}

\newcommand{\vct}[1]{\bm{#1}}
\newcommand{\mtx}[1]{\bm{#1}}

\definecolor{ejc}{RGB}{0,0,255}

\numberwithin{equation}{section} 

\def \endprf{\hfill {\vrule height6pt width6pt depth0pt}\medskip}

\title{Fitting ReLUs via SGD and Quantized SGD} 
\date{}
\author{Seyed~Mohammadreza~Mousavi~Kalan, Mahdi~Soltanolkotabi, and A.~Salman~Avestimehr\\ 
Ming Hsieh Department of Electrical Engineering, University of Southern California \\Email: mmousavi@usc.edu, soltanol@usc.edu, avestimehr@ee.usc.edu}

\begin{document}

\maketitle

\input{abstract}

\input{intro}

\input{Algorithms}

\input{main-results}

\input{Numerical}
\input{Related-work}

\input{Proofs.tex}

\input{Acknowledgement}

\bibliographystyle{IEEEtran}
\bibliography{iclr2019_conference}
\input{Appendix.tex}
\end{document}


%

%

\twocolumn[

\aistatstitle{Supplementary Material for \\Fitting ReLUs via SGD and Quantized SGD}

\aistatsauthor{ Anonymous Authors }

\vspace{5mm}
]



In the statement of the two Theorems of this paper, the probability of success depends on the proximity of the initial estimate to the solution.  To alleviate this dependency we directly adapt the \textit{Ensemble} Algorithm used in \cite{doi:10.1093/imaiai/iay005}. See Algorithm \ref{Alg1} for details of this approach.

\begin{algorithm}
    \caption{Ensemble Method}\label{algorithm1}
    \hspace*{\algorithmicindent} \textbf{Input:} Feature vectors $\vct{x}_1,\vct{x}_2,...,\vct{x}_n,$ labels $y_1,y_2,...,y_n$, relative error tolerance $\epsilon$, iteration count $K$ to achieve the desired error $\epsilon$, trial count $L$.\\
    \hspace*{\algorithmicindent} \textbf{Output:} An estimate $\vct{\hat{w}}$ for $\vct{w}^*.$ 
    \begin{algorithmic}[1]
    
   
    \State  Initialize $\vct{w}_0$ to satisfy the condition of Theorems 1 and 2.
    \State \textbf{for} $l=1,...,L$ run $K$ SGD updates from the initial point $\vct{w}_0$ to obtain $\vct{w}_K^{(l)}$.
    \For{$l=1,...,L,$}
    \If{$|B(\vct{w}_k^{(l)},2\sqrt{\epsilon})\cap \{\vct{w}_K^{(1),...,\vct{w}_K^{(L)}}\}|\geq L/2$} 
    \State \textbf{Return} $\vct{\hat{w}}:=\vct{w}_K^{(l)}$
    \end{algorithmic}
    \label{Alg1}
    \end{algorithm}
We now state a result regarding the performance of the ensemble method. We note that the proof of this lemma is essentially identical to its counterpart in \cite{doi:10.1093/imaiai/iay005} and requires only minor modifications.

\begin{proposition}{(Guarantees for ensemble methods).} Consider the setup and assumptions of Theorem 1. Furthermore assume that $\text{Prob}(\text{failure})\leq 1/3$. Then, for any $\delta'>0$ there is an absolute constant $C$ such that if $L\geq C\log{(1/\delta')}$ then the estimate $\vct{\hat{w}}$ obtained via Algorithm \ref{algorithm1} satisfies $||\vct{\hat{w}}-\vct{w}^*||_2\leq 9\epsilon||\vct{w}_0-\vct{w}^*||_2$.

\end{proposition}


\bibliography{iclr2019_conference}
\bibliographystyle{plain}

%% file: abstract.tex
\begin{abstract}
  In this paper we focus on the problem of finding the optimal weights of the shallowest of neural networks consisting of a single Rectified Linear Unit (ReLU). These functions are of the form $\vct{x}\rightarrow \max(0,\langle\vct{w},\vct{x}\rangle)$ with $\vct{w}\in\R^d$ denoting the weight vector. We focus on a planted model where the inputs are chosen i.i.d. from a Gaussian distribution and the
labels are generated according to a planted weight vector. We first show that mini-batch stochastic gradient descent when suitably initialized, converges at a geometric rate to the planted model with a number of
samples that is optimal up to numerical constants. Next we focus on a parallel implementation where in each iteration the mini-batch gradient is calculated in a distributed manner across multiple processors and then broadcast to a master or all other processors. To reduce the communication cost in this setting we utilize a Quanitzed Stochastic Gradient Scheme (QSGD) where the partial gradients are quantized. Perhaps unexpectedly, we show that QSGD maintains the fast convergence of SGD to a globally optimal model while significantly reducing the communication cost. We further corroborate our numerical findings via various experiments including distributed implementations over Amazon EC2.
\end{abstract}

%% file: intro.tex
\section{Introduction}

Many modern learning tasks involve fitting nonlinear models to data. Given training data consisting of $n$ pairs of input features $\vct{x}_i\in\R^d$ and desired outputs $y_i\in\R$ we wish to infer a function that best explains the training data. A prominent example is neural network models which have enabled impressive empirical success in applications spanning natural language processing to robotics. Guaranteed training of nonlinear data models however remain elusive. The main challenge is that fitting such nonlinear models requires solving highly nonconvex optimization problems and it is not clear why local search methods such as stochastic gradient descent converge to globally optimal solutions without getting stuck in spurious local optima and saddles.

In this paper we focus on fitting Rectified Linear Units (ReLUs) to the data which are functions $\phi_{\vct{w}}:\R^d\rightarrow\R$ of the form $\phi_{\vct{w}}(\vct{x})=\text{max}(0,\langle\vct{w},\vct{x}\rangle)$. We study a nonlinear least-squares formulation of the form
\begin{align}\label{lossfunction}
    \underset{\vct{w} \in \mathbb{R}^d}{\operatorname{min}}\ \mathcal{L}(\vct{w}):=&\frac{1}{n}\sum_{i=1}^{n}\ell_i(\vct{w})\nonumber\\=&\frac{1}{n}\sum_{i=1}^{n}(\text{max}(0,\langle\vct{w},\vct{x}_i\rangle)-y_i)^2.
\end{align}
A popular approach to solving problems of this kind is via Stochastic Gradient Descent (SGD). Indeed, SGD due to its manageable memory and footprint and highly parallelizable nature has become a mainstay of modern learning systems. Despite its wide use however, due to the nonconvex nature of the loss it is completely unclear why SGD converges to a globally optimal model. Fitting ReLUs via SGD poses new challenges: When are the iterates able to converge to global optima? How many samples are required? What is the convergence rate and is it possible to insure a fast, geometric rate of convergence? How do the answers to above change based on the mini-batch size? 

Yet, another challenge that arises when implementing SGD in a distributed framework is the high communication overhead required for transferring gradient updates between processors. A recent remedy is the use of quantized gradient updates such as Quantized SGD (QSGD) to reduce communication overhead. However, there is little understanding of how such quantization schemes perform on nonlinear learning tasks. Do quantized updates converge to the same solution as unquantized variants? If so, how is the convergence rate affected? How many quantization levels or bits are required to achieve good accuracy?


In this paper we wish to address the above challenges. Our main contributions are as follows:

\begin{itemize}
\item We study the problem of fitting ReLUs and show that SGD converges at a fast linear rate to a globally optimal solution. This holds with a near minimal number of data observations. We also characterize the convergence rate as a function of the SGD mini-batch sizes.
\item We show that the QSGD approach of \cite{NIPS2017_6768} also converges at a linear rate to a globally optimal solutions. This holds even when the number of quantization levels grows only Logarithmically in problem dimension. We also characterize the various tradeoffs between communication and computational resources when using such low-precision algorithms.

\item We provide experimental results corroborating our theoretical findings.

\end{itemize}

%% file: Algorithms.tex
\section{Algorithms: SGD and QSGD}
In this section we discuss the details of the algorithms we plan to study. We begin by discussing the specifics of the SGD iterates we will use. We then discuss how to use quantization techniques in order to reduce the communication overhead in a distributed implementation.

\subsection{Stochastic Gradient Descent (SGD) for fitting ReLUs}
To solve the optimization problem (\ref{lossfunction}) we use a mini-batch SGD scheme. While, the loss function (\ref{lossfunction}) is not differentiable, one can still use an update akin to SGD by defining a generalized notion of gradients for non-differentiable points as a limit of gradients of points converging to the non-differentiable point \cite{Clark}. Then, in each iteration we sample the indices $i_t^{(1)},i_t^{(2)},...,i_t^{(m)}$ uniformly with replacement from $\{1,2,\ldots,n\}$ and apply updates of the form
\begin{align}\label{sgd-updates}
    \vct{w}_{t+1}=\vct{w}_{t}-\eta\cdot\frac{1}{m}\sum_{j=1}^m \nabla\ell_{i_t^{(j)}}(\vct{w}_t).
\end{align}
Here, $\nabla \ell_{i}$ denotes the generalized gradient of the loss $\ell_i$ and is equal to
\begin{align*}
    \nabla \ell_i(\vct{w})=2\big(\text{ReLU}\left(\langle\vct{w},\vct{x}_i\rangle\right)-y_i\big)\big(1+\text{sgn}(\langle\vct{w},\vct{x}_i\rangle)\big)\vct{x}_i.
\end{align*}


\subsection{Reducing the communication overhead via Quantized SGD}\label{QSGDAlg}
One of the major advantages of SGD is that it is highly scalable to massive amounts of data. SGD can be easily implemented in a distributed platform where each processor calculates a portion of the mini-batch based on the available local data. Then the partial gradients are sent back to a master node or the other processors to calculate the full mini-batch gradient and update the next iteration. The latter case for example is common in modern deep learning implementations \cite{NIPS2017_6768}. Both distributed approaches however, suffer from a major bottleneck due to the cost of transmitting the gradients to the master or between the processors. 

A recent remedy for reducing this cost is utilizing lossy compression to quantize the gradients prior to transmission/broadcast. In particular, a recent paper \cite{NIPS2017_6768} proposes the quantized SGD (QSGD) algorithm based on a randomized quantization function $Q_s:\R^d\rightarrow\R^d$ with $s$ denoting the number of quantization levels. Specifically, for a vector $\vct{v}\in\R^d$ the $i$th entry of the quantization function $Q_s(\vct{v})$ is given by
\begin{align*}
    Q_s(v_i)=||\vct{v}||_2\cdot\text{sgn}(v_i)\cdot \xi_i(\vct{v},s),
\end{align*}
where $\xi_i(\vct{v},s)$'s are independent random variables defined as
\[   
\xi_i(\vct{v},s) = 
     \begin{cases}
       h/s &\text{with prob.}\ 1-(\frac{|v_i|s}{||\vct{v}||_2}-h).\\
       (h+1)/s &\text{otherwise.}\\
     \end{cases}
\]
Here, $h\in [0,s)$ is an integer such that $\frac{|v_i|}{||\vct{v}||_2}\in[h/s,(h+1)/s]$ and we follow the convention that $\text{sign}(0)=0$.



To see how this quantization scheme can be used in a distributed setting consider a master-worker distributed platform consisting of $K$ worker processors numbered $1,2,...,K$ and a master processor. To run SGD on this platform we partition the $n$ training data points into $K$ batches of size $\frac{n}{K}$ with each worker processor storing one of the batches. In each iteration, the master broadcasts the latest model to all the workers. Each worker then chooses $m_k$ points from local available data points randomly and computes a partial gradient based on the selected data points using the latest model received from the master and quantizes the resulting partial gradient. The workers then send the quantized stochastic gradients to the master. The master waits for all the quantized partial stochastic gradients from the workers and then updates the model using their average. As a result the aggregate effect of QSGD leads to updates of the form

\begin{align}\label{qsgd-updates}
    \vct{w}_{t+1}=\vct{w}_{t}-\eta\cdot \frac{1}{K}\sum_{k=1}^{K}Q_s\left(\nabla \left\{\frac{1}{m_k}\sum_{j=1}^{m_k} \ell_{i_t^{(j)}}(\vct{w}_t) \right\}\right).
\end{align}
Since only the quantized partial gradients are transimitted between the processors, QSGD significantly reduces the number of communicated bits.

%% file: main-results.tex
\section{Main results}

\subsection{SGD for fitting ReLUs}
In this section we discuss our results for convergence of the SGD iterates.
\begin{theorem}\label{theorem1} Let $\vct{w}^*$ be a fixed weight vector, and the feature vectors $\vct{x}_i \in \mathbb{R}^d$ be i.i.d. Gaussian random vectors distributed as $\mathcal{N}(\mathbf{0},\mtx{I})$ with the corresponding labels given by $y_i=\text{max}(0,\langle x_i,\vct{w}^*\rangle)$. Furthermore, assume 
\begin{itemize}
\item[(I)] the number of samples obey $n>c_0 d,$ for a fixed numerical constant $c_0$. 
\item[(II)] the initial estimate $\vct{w}_0$ obeys $||\vct{w_0}-\vct{w}^*||_2\leq\sqrt{\delta_1}\frac{7}{200}||\vct{w}^*||_2$ for some $0<\delta_1\leq 1/2$.
\end{itemize}
Then, the Stochastic Gradient Descent (SGD) updates in (\ref{sgd-updates}) with mini-batch size $m \in \mathbb{N}$ and learning rate $\eta=\frac{3}{4(\frac{9d}{m}+\frac{25}{16})}$ obey 
\begin{align}\label{ineq-4}
    \operatorname{\mathbb{E}}[\|\vct{w}_t-\vct{w}^*\|_2^2]\leq {\underbrace{\left(1-\frac{9}{16(\frac{9d}{m}+\frac{25}{16})}\right)}_{\text{convergence rate}=\rho}}^t||\vct{w}_0-\vct{w}^*||_2^2
\end{align}
with probability at least $1-\delta_1-2(n+2)e^{-\gamma d}-2(n+25)e^{-\gamma n}$. Furthermore, if $t\geq(\log(2/\epsilon)+\log(1/\delta_2))\frac{1}{1-\rho}$ \ for\ $0<\delta_2\leq1$, then
\begin{align}
||\vct{w}_t-\vct{w}^*||_2^2\leq\epsilon||\vct{w}_0-\vct{w}^*||_2^2,
\end{align}
holds with probability at least $1-\delta_1-\delta_2-2(n+2)e^{-\gamma d}-2(n+25)e^{-\gamma n}$.
\end{theorem}
\begin{remark}\label{init}
We note that Theorem \ref{theorem1} requires the initial estimate $\vct{w}_0$ to be sufficiently close to the planted model (i.e.~$\leq\sqrt{\delta_1}\frac{7}{200}||\vct{w}^*||_2$). Such an initial estimate can be easy obtained by using one full gradient descent step at zero \cite{NIPS2017_6796}. The required sample complexity for this initialization to be effective is on the order of $n\ge c\frac{d}{\delta_1^2}$ for $c$ a fixed numerical constant. For the purposes of this result we will use $\delta_1$ a small constant so that on the order of $n\gtrsim d$ samples are sufficient for this initialization.
\end{remark}

\begin{remark}
Theorem \ref{theorem1} shows that the SGD iterates (\ref{sgd-updates}) converge to a globally optimal solution at a geometric rate. Furthermore, the required number of samples for this convergence to occur is nearly minimal and is on the order of the number of parameters $d$.
\end{remark}

\begin{remark}
Theorem \ref{theorem1} also characterizes the influence of mini-batch size on the convergence rate, illustrating the trade-off between the computational load and the convergence speed.
\end{remark}

\begin{remark}\label{prob}
In Theorem \ref{theorem1}, the probability of success depends on the distance of initial point to the optimal solution. As we detail in the appendix we can use an ensemble algorithm to reduce the failure probability arbitrarily small without the need to start from an initial point that is very close to the optimal solution. 
\end{remark}
\begin{remark}\label{first-remark} We note that for $m=n$, the updates in (\ref{sgd-updates}) reduce to full gradient descent and the guarantee (\ref{ineq-4}) takes the form $||\vct{w}_t-\vct{w}^*||_2^2\leq\rho^t||\vct{w}_0-\vct{w}^*||_2^2$. In this special case our result recovers that of \cite{NIPS2017_6796} up to a constant factor.
\end{remark}

\subsection{Quantized SGD}

We next focus on providing gurantees for QSGD.
\begin{theorem}\label{theorem2}
Consider the same setting and assumptions as Theorem \ref{theorem1}. Furthermore, consider a parallel setting with $K$ worker processors and a master per Section \ref{QSGDAlg} and assume that each worker computes $\frac{m}{K}$ partial gradients in each iteration  (i.e.~$m_k=m/K$). We run QSGD over these processors via the iterative updates in (\ref{qsgd-updates}). Then

\begin{align}
    \operatorname{\mathbb{E}}[\|\vct{w}_t-\vct{w}^*\|_2^2]\leq {\underbrace{\left(1-\frac{9}{16\left(\left(1+\text{min}(\frac{d}{s^2},\frac{\sqrt{d}}{s})\right)(\frac{9d}{m}+\frac{25}{16})+\frac{25}{16}\right)}\right)}_{\text{convergence rate}=\alpha}}^t||\vct{w}_0-\vct{w}^*||_2^2
\end{align}

holds with probability at least $1-\delta_1-2(n+2)e^{-\gamma d}-2(n+25)e^{-\gamma n}$. Furthermore, if $t\geq(\log(2/\epsilon)+\log(1/\delta_2))\frac{1}{1-\alpha}$ \ for\ $0<\delta_2\leq1$ then
\begin{align}
||\vct{w}_t-\vct{w}^*||_2^2\leq\epsilon||\vct{w}_0-\vct{w}^*||_2^2
\end{align}

with probability at least $1-\delta_1-\delta_2-2(n+2)e^{-\gamma d}-2(n+25)e^{-\gamma n}$.

\end{theorem}

\begin{remark}
As mentioned in Remark \ref{init} of Theorem \ref{theorem1}, in order to address the initialization issue we can use one full Gradient Descent pass from zero to find an initializaiton obeying the conditions of this theorem.
\end{remark}

\begin{remark}
Similar to the results of Theorem \ref{theorem1}, Theorem \ref{theorem2} shows geometric convergence of the QSGD iterates (\ref{qsgd-updates}) with a near minimal number of samples ($n\gtrsim d$). Furthermore, it characterizes the effect of both mini-batch size and quantization levels on the convergence rate. Specifically, by increasing the quantization levels, the iterates (\ref{sgd-updates}) converge faster. Perhaps unexpected, by choosing the number of the bits to be on the order of ${\log{\sqrt{d}}}$ the iterates (\ref{sgd-updates}) and (\ref{qsgd-updates}) converge with the same rate up to a constant factor. This allows QSGD to significantly reduce the communication load while maintaining a computational effort comparable to SGD. 
\end{remark}



%% file: Numerical.tex
\section{Numerical results and experiments on Amazon EC2}
In this section we wish to investigate the results of Theorems \ref{theorem1} and \ref{theorem2} using numerical simulations and experiments on Amazon EC2. We first wish to investigate how the rate of convergence of mini-batch SGD and QSGD iterates depends on the different parameters. To this aim, we generate the planted weight vector $\vct{w}^* \in \mathbb{R}^d$ with $d=1000$ with entries distributed i.i.d. $ \sim N(200,3)$. In addition, we generate $n=10000$ feature vectors $\vct{x}_i\in \mathbb{R}^d$ i.i.d. $\sim N(0,1)$ and set the corresponding output labels $y_i=\text{max}(0,\langle \vct{x}_i,\vct{w}^*\rangle)$. To estimate $\vct{w}^*$, we start from a random initial point and run SGD and QSGD with learning rates $\eta=\frac{m}{d}$ and $\frac{m\cdot b}{9d}$, where $b$ is the number of bits required for quantization. 

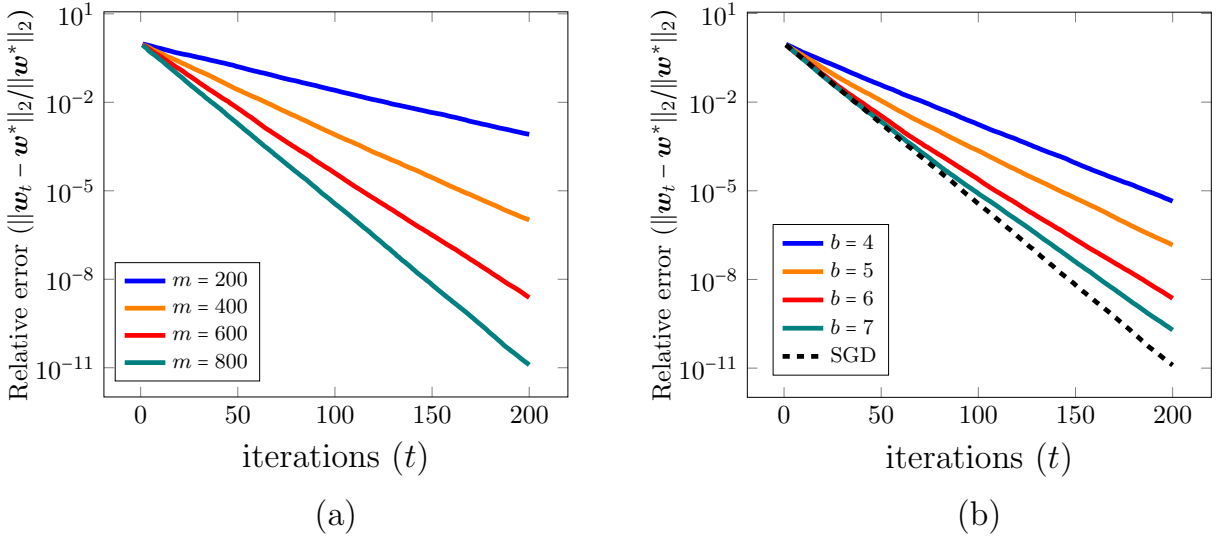
\begin{figure}
\centering
\hspace{-4cm}
    \begin{subfigure}{0.23\textwidth}
    \centering
    \begin{tikzpicture}[scale=.9]
    \begin{semilogyaxis}[xlabel=\huge$\substack{\text{iterations ($t$)}\\ \\ \text{(a)}}$\normalsize,
        ylabel=Relative error ($||\vct{w}_t-\vct{w}^*||_2/||\vct{w}^*||_2)$, legend style={font=\footnotesize,at={(0.18,0.35)},anchor=north,legend cell align=left}]

      \addplot [blue,line width=2pt] table[x index=0,y index=1]{m200.dat};\addlegendentry{$m=200$}
      
        \addplot [orange,line width=2pt] table[x index=0,y index=1]{m400.dat};\addlegendentry{$m=400$}
        
       \addplot [red,line width=2pt] table[x index=0,y index=1]{m600.dat};\addlegendentry{$m=600$}
       
      \addplot [teal,line width=2pt] table[x index=0,y index=1]{m800.dat};\addlegendentry{$m=800$}    
\end{semilogyaxis}
\end{tikzpicture}

 \centering
            \label{fig1a}
        \end{subfigure} 
        \hspace{4.5cm}
    \begin{subfigure}{0.23\textwidth}
    \centering
    \begin{tikzpicture}[scale=.9]



 \begin{semilogyaxis}[xlabel=\huge$\substack{\text{iterations ($t$)}\\ \\ \text{(b)}}$\normalsize,
        ylabel=Relative error ($||\vct{w}_t-\vct{w}^*||_2/||\vct{w}^*||_2)$, legend style={font=\footnotesize,at={(0.18,0.45)},anchor=north,legend cell align=left}]

      \addplot [blue,line width=2pt] table[x index=0,y index=1]{s4.dat};\addlegendentry{$b=4$}
      
        \addplot [orange,line width=2pt] table[x index=0,y index=1]{s5.dat};\addlegendentry{$b=5$}
        
       \addplot [red,line width=2pt] table[x index=0,y index=1]{s6.dat};\addlegendentry{$b=6$}
       
      \addplot [teal,line width=2pt] table[x index=0,y index=1]{s7.dat};\addlegendentry{$b=7$}    
      \addplot [dashed,line width=2pt] table[x index=0,y index=1]{m800.dat};\addlegendentry{SGD} 
\end{semilogyaxis}
\end{tikzpicture}


            \label{fig1b}
        \end{subfigure}

    \caption{ (a) This plot depicts the convergence behavior of mini-batch SGD iterates for various mini-batch sizes $m$.(b) This plot depicts the convergence behavior of QSGD iterates for various bits of quantization $b$ with the mini-batch size fixed at $m=800$.}

    \label{fig1}
    \end{figure}

In Figure \ref{fig1}\textcolor{darkred}{(a)} we focus on corroborating our convergence analysis for SGD. To this aim we vary the mini-batch size $m$ and plot the relative error ($\|\vct{w}-\vct{w}^*\|_{2}/\|\vct{w}^*\|_2$) as a function of the iterations. This figure demonstates that the convergence rate is indeed linear and increasing the batch size $m$ results in a faster convergence. 

In Figure \ref{fig1}\textcolor{darkred}{(b)} we focus on understanding the effect of the number of bits on the convergence behavior of QSGD. To this aim we fix the mini-batch size at $m=800$ and vary the number of bits $b$. This plot confirms that QSGD maintains a linear convergence rate comparable to SGD (especially when $b=7$).

\begin{figure}[http]
        \begin{subfigure}[!t]{\textwidth}
  \begin{center}
 \includegraphics[scale=0.1]{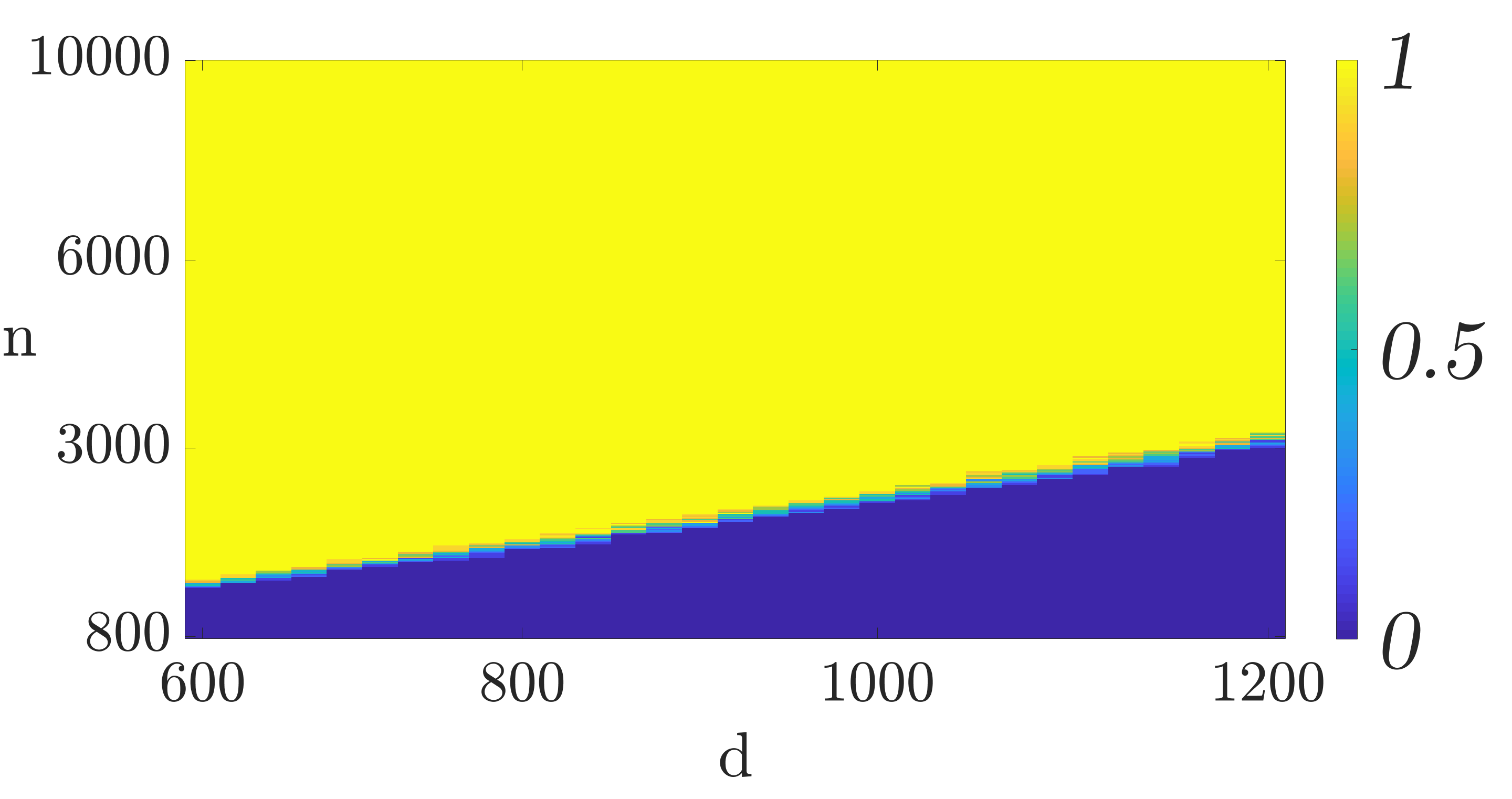}
 \end{center}
                \caption{}
                \label{6T_cell}
        \end{subfigure}\\
        \begin{subfigure}[!t]{\textwidth}
                \begin{center}
 \includegraphics[scale=0.1]{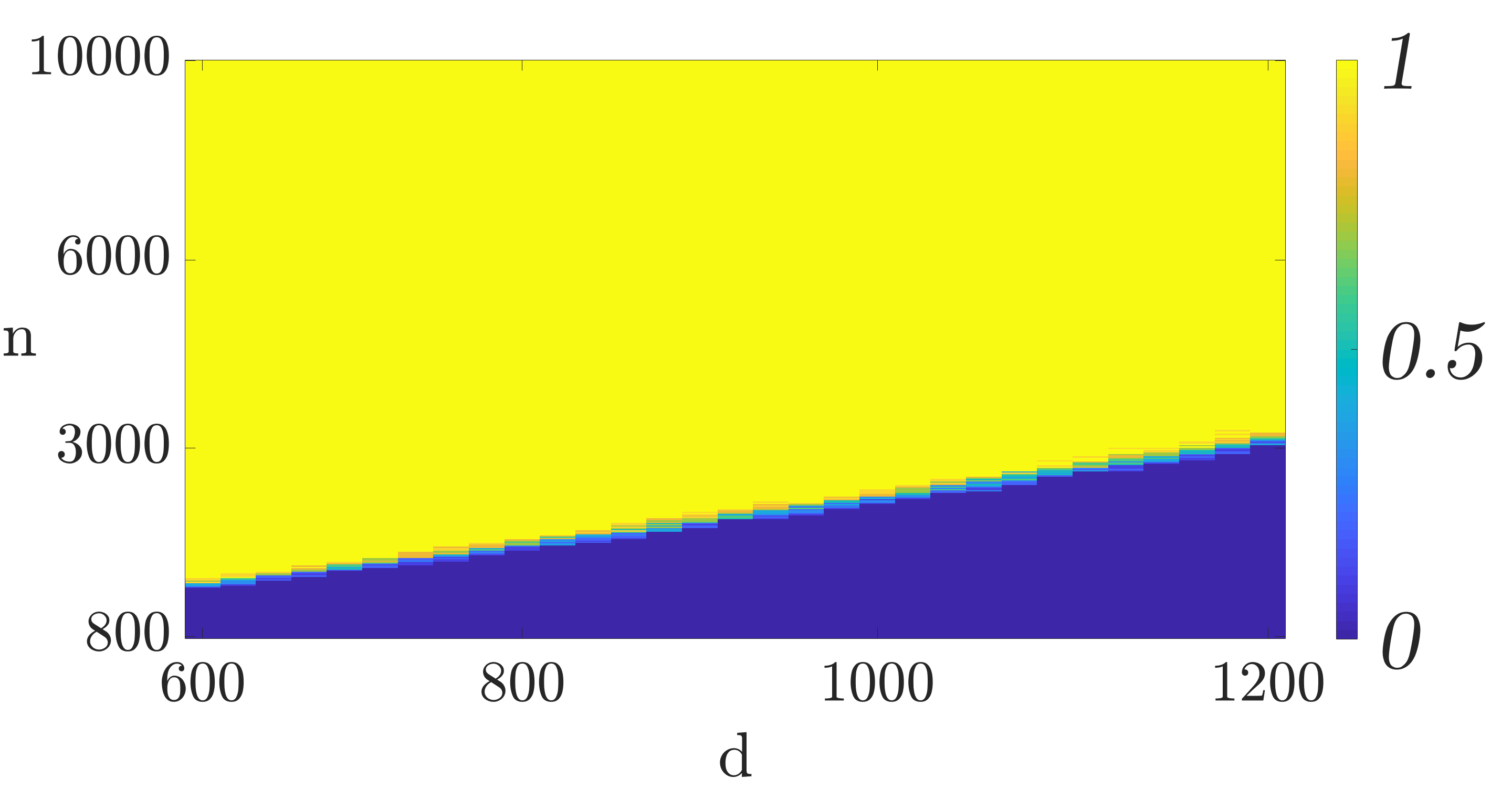}
 \end{center}
                \caption{}
                \label{6T_Layout}
        \end{subfigure}
        
        \caption{Empirical probability that (a) SGD and (b) QSGD with $b=7$ finds the global optimum for different number of data points ($n$) and feature dimensions ($d$).}
        \label{Splitter}
\end{figure}

To understand the required sample complexity of the algorithms, we plot phase transition curves depicting the empirical probability that gradient descent converges to $\vct{w}^*$ for different data set sizes ($n$) and feature dimensions ($d$). For each value of $n$ and $d$ we perform $10$ trials with the data generated according to the model discussed above. In each trial we run the algorithm for $2000$ iterations. If the relative error after $2000$ iterations is less than $10^{-3}$ we consider the trial a success otherwise a failure. Figures \ref{Splitter}\textcolor{darkred}{(a)} and \ref{Splitter}\textcolor{darkred}{(b)} depict the phase transition for mini-batch SGD and QSGD with $b=7$ bits, respectively. As it can be seen, the figures corroborate the linear relationship of the required sample complexity with the feature dimensions. Furthermore, these figures demonstrate that quantization does not substantially change the required sample complexity.

In order to understand the effectiveness of QSGD at reducing the communication overhead, we provide experiments on Amazon EC2 clusters and compare the performance of QSGD with SGD. We utilize a master-worker architecture and make use of \textbf{t2.micro} instances. We use Python for implementing two algorithms and use MPI4py \cite{dalcin2011parallel} for message passing across the instances. Before starting the iterative updates each worker receives its portion of the training data. In each iteration $t$, after the workers receive the latest model $\vct{w}_t$ from the master, they compute the stochastic gradients at $\vct{w}_t$ based on the local data and then send it back to the master. In SGD and QSGD we use \textit{float64} and \textit{int8} for sending gradients, respectively. Additionally, we use \texttt{isend()} and \texttt{irecv()} for sending and receiving, respectively, and \texttt{Time.time()} to measure the running time.

We compare the performances of these two algorithms in the following two scenarios.

\begin{figure}[htbp]
  \centering
  \includegraphics[height =.45\textwidth,width=.8\textwidth]{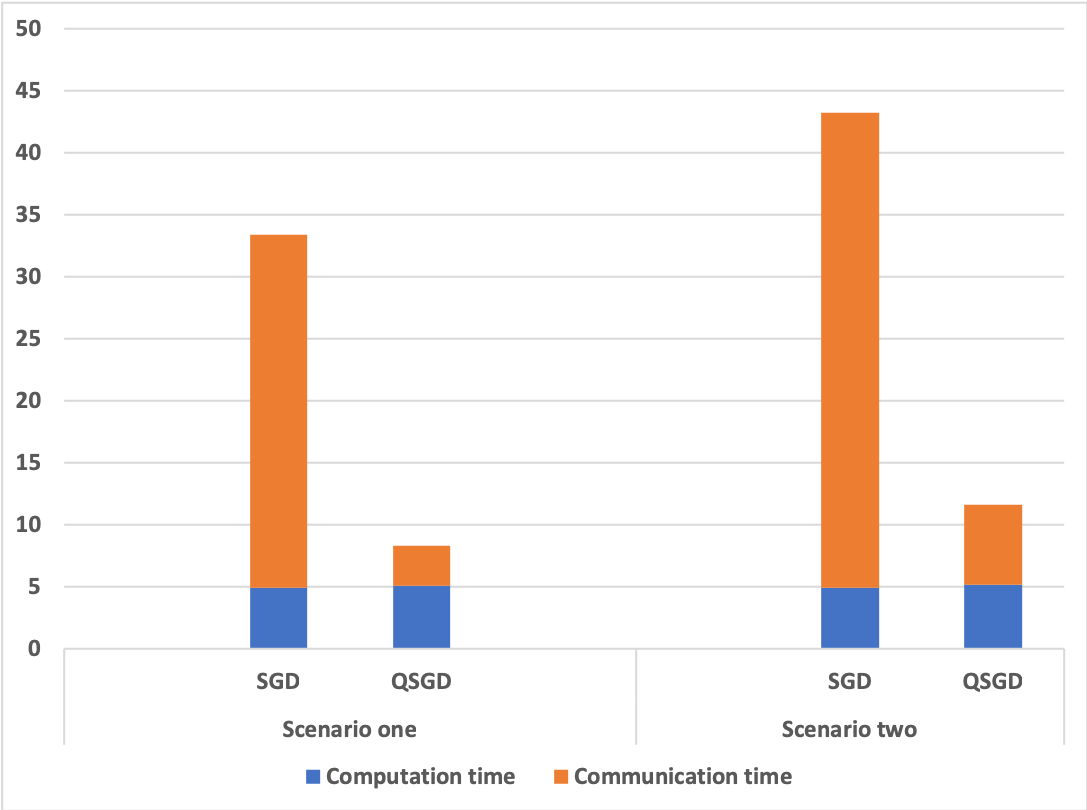}
  \caption{Run-time compassion of SGD and QSGD on Amazon EC2 clusters for the two scenarios.}
  \label{fig:run-time1}
 
\end{figure}

\begin{itemize}
\item Scenario one: We use 41 \textbf{t2.micro} instances, with one master and $40$ workers. We have $n=20000$ data points with feature dimension $d=4000$. We partition and distribute the data into $40$ equal batches of size $20000/40=500$ with each worker performing updates based its own batch.
\item Scenario two: We use 51 \textbf{t2.micro} instances, with one master and $50$ workers. In this scenario we use $n=25000$ and $d=4000$, and again distribute the data evenly among the workers.
\end{itemize}
Table \ref{table:scenarios} summarizes the experiment scenarios. In both cases we run SGD and QSGD algorithms for $300$ iterations. Figure \ref{fig:run-time1} depicts the total running times. Table \ref{break} also shows the breakdowns of the run-times. These experiments indicate that the total running time of QSGD is 5 times less than that of SGD. Since both algorithms have similar convergence rate and hence computational time, this clearly demonstrates that the communication time for QSGD is significantly smaller.

\begin{table}[htbp]

\caption{Experiment scenarios.}
\label{table:scenarios}
  \centering
  \scalebox{0.9}{
  \begin{tabular}{| c | c | c | c | }
    \hline
    scenario index & \# of workers ($K$)& \# of data points ($n$)& feature dimension ($d$) \\ \hline
    1 & 40 & 20000  &4000  \\ \hline
    2 & 50 & 25000  &4000  \\ \hline
   
  \end{tabular}}
\end{table}

\begin{table}[!htbp]

 \caption{Breakdowns of the run-times in the both scenarios.}
 \vspace{0.5mm}
\label{table:scenario one}
  \centering
  \scalebox{1}{
  \begin{tabular}{|c | c | c| c | c |}
    \hline
    schemes & scenario index& comm. time &comp. time & total time \\
     
     \hline
    SGD & 1 & 28.5100 s &4.921 s &  33.431 s\\ \hline
    QSGD & 1 &3.2470 s  & 5.056 s  &8.303 s\\ \hline
    SGD & 2 &38.2910 s  & 4.94 s  &43.231 s\\ \hline
    QSGD & 2 &6.5010 s  & 5.169 s  &11.67 s\\ \hline
   
  \end{tabular}}
  \label{break}
\end{table}

%% file: Related-work.tex
\vspace{-.4cm}
\section{Related work}
The problem of fitting non-linear models to data has a rich history in statistics and learning theory using a variety of algorithms \cite{conf/colt/KalaiS09,DBLP:journals/corr/GoelKKT16,bb24efcb23e343d48a4902f9e2dd1a7b,RePEc:fth:minner:264}. In the context of training deep models such nonlinear learning problems have lead to major breakthroughs in various applications \cite{Krizhevsky:2012:ICD:2999134.2999257,Collobert:2008:UAN:1390156.1390177}. Despite these theoretical and empirical advances, rigorous understanding of local search heuristics such as stochastic gradient descent which are arguably the most widely used techniques in practice, have remained elusive.

Recently, there has been a surge of activity surrounding nonlinear data fitting via local search. Focusing on ReLU nonlinearities \cite{NIPS2017_6796} shows that in generic instances full gradient descent converges to a globally optimal model at a geometric rate. See also \cite{oymak2018stochastic,wang2018learning,jagatap2018learning,zhang2018learning,liang2018understanding,goel2017learning,goel2018learning,fu2018local} for a variety of related theoretical and empirical work for fitting ReLUs and shallow neural networks via local search heuristics. In contrast to the above, which require full gradient updates in this paper we focus on fitting ReLU nonlinearities via mini-batch SGD and QSGD. Finally, we would like to mention another recent result which studies a stochastic method for fitting ReLUs \cite{DBLP:journals/corr/abs-1803-09357} via a First order Perturbed Stochastic Gradient Descent. This approach differs from ours in a variety of ways including the update strategy and required sample complexity. In particular, this approach is based on stochastic methods with random search strategies which only use function evaluations. Furthermore, this result requires $n\gtrsim d^4$ samples and provides a polynomial converge guarantee where as is this paper we have established a geometric rate of convergence to the global optima with a near minimal number of samples that scales linearly in the problem dimension (i.e.~ $n\gtrsim d$).

Recently there has been a lot of exciting activity surrounding SGD analysis. Classic SGD convergence analysis shows that the distance to the global optima (in loss or parameter value) decreases polynomially in the number of iterations (e.g.~$1/t$ after $t$ iterations). More recent results demonstrate that in certain cases, a significantly faster and geometric rate of convergence (i.e. $\rho^{-t}$ with $\rho<1$) is possible
\cite{NIPS2014_5355,NIPS2011_4316}. For instance, \cite{Strohmer2008} showed that randomized Kaczmarz algorithm converges to the solution of a consistent linear system of equations at a geometric rate. More recently, \cite{DBLP:journals/corr/abs-1712-06559}  showed that under some assumptions such as smoothness, strong convexity, and perfect interpolation, SGD achieves a geometric rate of convergence. In this paper we add to this growing literature and demonstrate that such a fast geometric rate of convergence to a globally optimal solution is also possible despite the nonlinear and nonconvex nature of fitting ReLUs.

Reducing communication overhead by compressing the gradients has become increasing popular in recent literature \cite{yu2018gradiveq,NIPS2015_5717,NIPS2012_4687}. Most notably \cite{1-bit-stochastic-gradient-descent-and-application-to-data-parallel-distributed-training-of-speech-dnns} empirically demonstrated that one-bit quantization of gradients is highly effective for scalable training of deep models. The QSGD paper \cite{NIPS2017_6768} develops convergence gurantees for convex losses as well convergence of gradient to zero for nonconvex losses. Related \cite{DBLP:journals/corr/abs-1802-04434} also shows that under the assumption of smoothness and bounded variance a quantized SGD procedure converges to a stationary point of a general non-convex function with a polynomial convergence rate. In contrast in this paper we focus on geometric convergence to the global optima but for the specific problem of fitting a ReLU  nonlinearity.






%% file: Proofs.tex
\section{Proofs}
\subsection{Convergence analysis for fitting ReLUs via SGD (Proof of Theorem \ref{theorem1})}

To show
\begin{align}
\label{thm1conc}
    \underset{\vct{w}_t}{\operatorname{\mathbb{E}}}\left[||\vct{w}_t-\vct{w}^*||_2^2\right]\leq\rho||\vct{w}_{t-1}-\vct{w}^*||_2^2,
\end{align}
we begin with a useful definition. In particular, we denote the mini-batch empirical loss as $L_{I_{m}(t)}(\vct{w})\delequal \frac{1}{m}\sum_{j=1}^{m}\ell_{i_t^{(j)}}(\vct{w})$, where $I_{m}(t)=\{i_t^{(1)},i_t^{(2)},...,i_t^{(m)}\}$ is the set of indices chosen at iteration $t$. Therefore, we can rewrite the mini-batch SGD updates as
\begin{align*}
    \vct{w}_t-\vct{w}^*=\vct{w}_{t-1}-\vct{w}^*-\eta\nabla L_{I_{m}(t)}(\vct{w}_{t-1}).
\end{align*}
To upper bound $\mathbb{E}||\vct{w}_t-\vct{w}^*||_2^2$ in terms of $\mathbb{E}||\vct{w}_{t-1}-\vct{w}^*||_2^2$, we expand $\mathbb{E}||\vct{w}_t-\vct{w}^*||_2^2$ in the form
\begin{align}\label{expandmain}
    \mathbb{E}_{I_{m}(t)}||\vct{w}_t-\vct{w}^*||_2^2=&\mathbb{E}_{I_{m}(t)}\bigg[||\vct{w}_{t-1}-\vct{w}^*||_2^2\bigg]\\&+\mathbb{E}_{I_{m}(t)}\bigg[-2\eta\langle\vct{w}_{t-1}-\vct{w}^*,\nabla L_{I_{m}(t)}(w_{t-1})\rangle\nonumber+\eta^2||\nabla L_{I_{m}(t)}(w_{t-1})||_2^2\bigg]
\end{align}
To draw the desired conclusion of the theorem (i.e.~(\ref{thm1conc})), it suffices to upper bound the second expectation which consists of two terms
. To upper bound the first term, We apply the expectation to the inner product to conclude that
\begin{align}\label{tmpthm1}
        \mathbb{E}_{I_{m}(t)}&\left[\langle\vct{w}_{t-1}-\vct{w}^*,\nabla L_{I_{m}(t)}(w_{t-1})\rangle\right]=\langle\vct{w}_{t-1}-\vct{w}^*,\nabla\mathcal{L}(\vct{w}_{t-1})\rangle
    \end{align}
To lower bound the inner product we utilize the following lemma proved in \cite{NIPS2017_6796}.
\begin{lemma}\label{RELULEMMA}
For the loss function defined in (\ref{lossfunction}), we have
\begin{align*}
&\langle \vct{u},\vct{w}-\vct{w}^*-\nabla \mathcal{L}(\vct{w})\rangle\ \leq 2\left(\delta+\sqrt{1+\delta}\left(1+\frac{2}{(1-\epsilon)^2}\right)\left(\delta +\sqrt{\frac{21}{20}\epsilon}\right)\right)\cdot||\vct{w}-\vct{w}^*||_2,
\end{align*}
holding for all $u \in \mathcal{B}^d$ and $w \in E(\epsilon)=\left\{\vct{w}\in \mathbb{R}^d: ||\vct{w}-\vct{w}^*||_2\leq \epsilon ||\vct{w}^*||_2\right\}$ with probability at least $1-16e^{-\gamma \delta^2 n}-(n+10)e^{-\gamma n}$. Specifically for $\delta=10^{-4}$ and $\epsilon=7/200$, we obtain
\begin{align}\label{ineq}
    \langle\vct{u},\vct{w}_{t-1}-\vct{w}^*-\nabla \mathcal{L}(\vct{w}_{t-1})\rangle\leq\frac{1}{4}||\vct{w}_{t-1}-\vct{w}^*||_2.
    \end{align}
\end{lemma}
Using Lemma \ref{RELULEMMA} we have
\begin{align*}
    \langle\vct{u},\vct{w}_{t-1}-\vct{w}^*-\nabla \mathcal{L}(\vct{w}_{t-1})\rangle\leq\frac{1}{4}||\vct{w}_{t-1}-\vct{w}^*||_2
    \end{align*}
for all $\vct{u}$ with $||\vct{u}||_2\leq1$. By choosing $u=\frac{\vct{w}_{t-1}-\vct{w}^*}{||\vct{w}_{t-1}-\vct{w}^*||_2}$ we can conclude that
\begin{align}\label{inner}
        \langle \vct{w}_{t-1}-\vct{w}^*,\nabla \mathcal{L}(w_{t-1})\rangle\geq\frac{3}{4}||\vct{w}_{t-1}-\vct{w}^*||_2^2
    \end{align}
holds with probability at least $1-16e^{-10^{-8}\gamma n}-(n+10)e^{-\gamma n}$.

Thus, by combining (\ref{expandmain}), (\ref{tmpthm1}), and (\ref{inner}) we can conclude that
\begin{align}\label{expand2}
    \mathbb{E}_{I_{m}(t)}||\vct{w}_t-\vct{w}^*||_2^2\leq&(1-\frac{3\eta}{2})||\vct{w}_{t-1}-\vct{w}^*||_2^2+\eta^2\mathbb{E}_{I_{m}(t)}||\nabla L_{I_{m}(t)}(\vct{w}_{t-1})||_2^2.
\end{align}
Next, in order to upper bound the second term in the right hand side of (\ref{expand2}), we use following lemma proven in Section \ref{seclemsec}.
\begin{lemma}\label{second-moment}
The following inequality
 \begin{align}\label{bound}
    \mathbb{E}_{I_{m}(t)}||\nabla L_{I_{m}(t)}(\vct{w}_{t-1})||_2^2 \leq&\left(\frac{9d}{m}+\frac{25}{16}\right)||\vct{w}_{t-1}-\vct{w}^*||_2^2
 \end{align}
holds with probability at least $1-2(n+1)e^{-\gamma d}-(n+25)e^{-\gamma n}$.
\end{lemma}
Hence, by (\ref{expand2}) and (\ref{bound}) with $\eta^*=\frac{3}{4(\frac{9d}{m}+\frac{25}{16})}$ we arrive at 
\begin{align}\label{last-ineq}
     \mathbb{E}_{I_{m}(t)}||\vct{w}_t-\vct{w}^*||_2^2&\leq
     \bigg[1-\frac{3}{2}\eta+(\frac{9d}{m}+\frac{25}{16})\eta^2\bigg]||\vct{w}_{t-1}-\vct{w}^*||_2^2\nonumber\\
     &\le(1-\frac{9}{16(\frac{9d}{m}+\frac{25}{16})})\mathbb{E}||\vct{w}_{t-1}-\vct{w}^*||_2^2\nonumber\\
     &=\rho\cdot\mathbb{E}||\vct{w}_{t-1}-\vct{w}^*||_2^2.
    \end{align}
This holds with probability at least $1-2(n+1)e^{-\gamma d}-2(n+25)e^{-\gamma n}$. We note however, that the proof is not yet complete. We have not shown that all subsequent iterations will also lie in the local neighborhood dictated by Lemma \ref{RELULEMMA}. We have only shown that on \emph{average} they belong to this neighborhood. To overcome this challenge we utilize some techniques from stochastic processes to obtain a conditional convergence. Our argument heavily borrows from \cite{doi:10.1093/imaiai/iay005} with some parts directly adapted.

\textit{Conditional linear convergence}. In order to make our notations compatible with typical notations used in stochastic processes theory, we denote $\vct{W}_k$ as the random vector of estimated solution at iteration $k$ and $\vct{w}_k$ as a realization of that random vector. Using these conventions the result we have proven so far can be rewritten as
    \begin{align*}
    \mathbb{E}[||\vct{W}_{k+1}-\vct{w}^*||_2^2|\vct{W}_k=\vct{w}_k]\leq\rho||\vct{w}_k-\vct{x}||_2^2.
    \end{align*}
    Let $\mathcal{F}_k$ denote the $\sigma$-algebra generated by indices chosen in the steps from $1$ to $k$. Also let $B \subset R^d$ be the region consisting of all points which are in $E(\epsilon)=\left\{\vct{w}\in \mathbb{R}^d: ||\vct{w}-\vct{w}^*||_2\leq \epsilon ||\vct{w}^*||_2\right\}$ where $\epsilon=\frac{7}{200}$. Finally, assume an initial estimate which is fixed obeying $\vct{w}_0 \in B$ and $||\vct{w}_0-\vct{w}^*||_2\leq\sqrt{\delta_1}\epsilon||\vct{w}^*||_2$. Now define a stopping time $\tau$ as $\tau:=\text{min}\{k: \vct{W}_k\notin B\}$. For each $k$, and $\vct{w}_k \in B$, we have
    \begin{align*}
        \mathbb{E}[||\vct{W}_{k+1}-\vct{w}^*||_2^2 1_{\tau>k+1}|\vct{W}_k=\vct{w}_k]\leq&\mathbb{E}[||\vct{W}_{k+1}-\vct{w}^*||^2 1_{\tau>k}|\vct{W}_k=\vct{w}_k]\nonumber\\
        =& \mathbb{E}[||\vct{W}_{k+1}-\vct{w}^*||_2^2 1_{\tau>k}|\vct{W}_k=\vct{w}_k,\mathcal{F}_k]\nonumber\\
        =& \mathbb{E}[||\vct{W}_{k+1}-\vct{w}^*||_2^2 |\vct{W}_k=\vct{w}_k,\mathcal{F}_k]1_{\tau>k}\nonumber\\
        \leq & \rho||\vct{w}_k-\vct{w}^*||_2^21_{\tau>k}
    \end{align*}
    Hence
    $$\mathbb{E}[||\vct{W}_{k+1}-\vct{w}^*||_2^2 1_{\tau>k+1}]\leq\rho\mathbb{E}[||\vct{W}_{k}-\vct{w}^*||_2^21_{\tau>k}]$$
    Therefore, we obtain
    \begin{align}\label{infty}
    \mathbb{E}[||\vct{W}_{k}-\vct{w}^*||_2^2 1_{\tau>k}]\leq\rho^k||\vct{w}_0-\vct{w}^*||_2^2.
    \end{align}

 To show that the probability of leaving this neighborhood is low we bound $\mathbb{P}(\tau<\infty)$. We begin by showing that $Z_k=\frac{||\vct{W}_{\tau \wedge k}-\vct{w}^*||_2^2}{\rho^{\tau \wedge k}}$ is a supermartingale.
   \begin{align*}\mathbb{E}[Z_{k+1}|\mathcal{F}_k]=& \mathbb{E}[\frac{||\vct{W}_{\tau \wedge (k+1)}-\vct{w}^*||_2^2}{\rho^{\tau \wedge (k+1)}}1_{\{\tau\leq k\}}|\mathcal{F}_k]+\mathbb{E}[\frac{||\vct{W}_{\tau \wedge (k+1)}-\vct{w}^*||_2^2}{\rho^{\tau \wedge (k+1)}}1_{\{\tau> k\}}|\mathcal{F}_k]\nonumber\\
=& \mathbb{E}[\frac{||\vct{W}_{\tau \wedge (k)}-\vct{w}^*||_2^2}{\rho^{\tau \wedge (k)}}1_{\{\tau\leq k\}}|\mathcal{F}_k]+\mathbb{E}[\frac{||\vct{W}_{k+1}-\vct{w}^*||_2^2}{\rho^{k+1}}1_{\{\tau> k\}}|\mathcal{F}_k]\nonumber\\
\leq& Z_k1_{\{\tau \leq k\}}+\rho \frac{1}{\rho^{k+1}}\mathbb{E}[||\vct{W}_k-\vct{w}^*||_2^2|\mathcal{F}_k]1_{\{\tau>k\}}\nonumber\\
=& Z_k1_{\{\tau \leq k\}}+Z_k1_{\{\tau > k\}}\nonumber\\
=&Z_k.
\end{align*}
Using the fact that $Z_k$ is a supermartingale we have
\begin{align*}
   Z_0\geq& \mathbb{E}[Z_k|\mathcal{F}_0]\\\geq&\mathbb{E}[\frac{||\vct{W}_{k \wedge \tau}-\vct{w}^*||_2^2}{\rho^{k \wedge \tau}}1_{k\geq \tau}|\mathcal{F}_0]\nonumber\\
   \geq& \mathbb{E}[\frac{||\vct{W}_{\tau}-\vct{w}^*||_2^2}{\rho^{\tau}}1_{k\geq \tau}|\mathcal{F}_0].
\end{align*}
Now By the definition of stopping time,
\begin{align*}
  ||\vct{W}_{\tau}-\vct{w}^*||_2^2\geq \epsilon^2 ||\vct{w}^*||_2^2
\end{align*}
and using $||\vct{w}_0-\vct{w}^*||_2\leq \epsilon\sqrt{\delta_1}||\vct{w}^*||_2$, we arrive at
\begin{align*}
\epsilon^2\delta_1||\vct{w}^*||_2^2\geq \mathbb{E}[\frac{\epsilon^2||\vct{w}^*||_2^2}{\rho^{\tau}}1_{\{k\geq \tau\}}|\mathcal{F}_0].
\end{align*}
This in turn implies that $\delta_1\geq \mathbb{E}[\frac{1_{\{k\geq \tau\}}}{\rho^{\tau}}|\mathcal{F}_0]$. Thus, 
\begin{align*}
\delta_1\geq \mathbb{E}[\frac{1_{\{k\geq \tau\}}}{\rho^{\tau}}|\mathcal{F}_0]
\geq&\mathbb{E}[1_{\{k\geq \tau\}}|\mathcal{F}_0]
=\mathbb{P}\{k\geq \tau\}.
\end{align*}
Whence,
\begin{align*}
 \delta_1=\lim_{k\to\infty} \delta_1\geq \lim_{k\to\infty}\mathbb{P}\{k\geq \tau\}=\mathbb{P}\{\infty>\tau\}.
\end{align*}   
Hence we conclude that $\mathbb{P}\{\tau<\infty\}\leq\delta_1\leq 1/2$. Thus
\begin{align*}
    \mathbb{E}[||\vct{W}_t-\vct{w}^*||_2^21_{\tau=\infty}]=&\mathbb{E}[||\vct{W}_t-\vct{w}^*||_2^2|\tau=\infty]\mathbb{P}(\tau=\infty)+0\cdot\mathbb{P}(\tau<\infty)\\
    \geq& \frac{1}{2}\mathbb{E}[||\vct{W}_t-\vct{w}^*||_2^2|\tau=\infty].
\end{align*}
By (\ref{infty}) we have
\begin{align*}
    \mathbb{E}[||\vct{W}_t-\vct{w}^*||_2^2|\tau=\infty]\leq 2\rho^t||\vct{w}_0-\vct{w}^*||_2^2.
\end{align*}
Thus, using Markov's inequality we conclude that
\begin{align*}
    \mathbb{P}(||\vct{W}_t-\vct{w}^*||_2^2>\epsilon||\vct{w}_0-\vct{w}^*||_2^2|\tau=\infty)&\leq\frac{\mathbb{E}[||\vct{W}_t-\vct{w}^*||_2^2|\tau=\infty]}{\epsilon||\vct{w}_0-\vct{w}^*||_2^2}\\
    &\leq \frac{2\rho^t}{\epsilon},
\end{align*}   
which is bounded by $\delta_2$.
\subsubsection{Proof of Lemma \ref{second-moment}}\label{seclemsec}
We use the following identity shown in \cite{DBLP:journals/corr/abs-1712-06559}.
\begin{align}\label{expanison-id}
    \mathbb{E}_{I_{m}(t)}||\nabla L_{I_{m}(t)}(\vct{w}_{t-1})||_2^2=&\frac{m-1}{m}||\nabla \mathcal{L}(\vct{w}_{t-1})||_2^2 +\frac{1}{m}\mathbb{E}_{I_{1}(t)}\left[||\nabla L_{I_{1}(t)}(\vct{w}_{t-1})||_2^2\right]
\end{align}
To upper bound the first term, using (\ref{ineq}) with $\vct{u}=\frac{-\nabla \mathcal{L}(\vct{w}_{t-1})}{||\nabla \mathcal{L}(\vct{w}_{t-1})||_2}$ we conclude that
\begin{align*}\label{first}
    ||\nabla \mathcal{L}(\vct{w}_{t-1})||_2\leq& \langle\frac{\nabla \mathcal{L}(\vct{w}_{t-1})}{||\nabla \mathcal{L}(\vct{w}_{t-1})||_2},\vct{w}_{t-1}-\vct{w}^*\rangle+\frac{1}{4}||\vct{w}_{t-1}-w^*||_2\nonumber\\
    \leq&||\vct{w}_{t-1}-\vct{w}^*||_2+\frac{1}{4}||\vct{w}_{t-1}-w^*||_2\nonumber\\
    =&\frac{5}{4}||\vct{w}_{t-1}-w^*||_2
\end{align*}
holds with probability at least $1-(n+25)e^{-\gamma n}$.

In order to upper bound the second term on the right hand side of (\ref{expanison-id}) we use the following chain of inequalities
\begin{align*}
    \mathbb{E}_{I_{1}(t)}\big[||\nabla L_{I_{1}(t)}(\vct{w}_{t-1})||_2^2\big]\nonumber=&\frac{1}{n}\sum_{i=1}^{n}||\big(\text{ReLU}(\langle\vct{w}_{t-1},x_i\rangle)-\text{ReLU}(\langle\vct{w}^*,x_i\rangle)\big)||_2^2\cdot(1+\text{sgn}\big(\langle\vct{w}_{t-1},x_i\rangle)\big)^2||x_i||_{2}^2\nonumber\\
    \leq&\frac{4}{n}\sum_{i=1}^{n}||x_i||_{2}^2|\langle\vct{w}_{t-1}-\vct{w}^*,\vct{x}_i\rangle|^2\nonumber\\
    \stackrel{a}{\le}&4\cdot \frac{3d}{2}(\vct{w}_{t-1}-\vct{w}^*)^T\frac{1}{n}\sum_{i=1}^{n}x_ix_i^T(\vct{w}_{t-1}-\vct{w}^*)\nonumber\\
    \stackrel{b}{\le}&4d\cdot\frac{3d}{2}\cdot\frac{3}{2}||\vct{w}_{t-1}-\vct{w}^*||_2^2\nonumber\\
    =&9d ||\vct{w}_{t-1}-\vct{w}^*||_2^2
    \end{align*}
This holds with probability at least $1-2(n+1)e^{-\gamma d}$. In (a) we used the fact that a high-dimensional Gaussian random vector is well-concentrated on the sphere ofradius $\sqrt{d}$ with high probability and in the last inequality we used a well-known upper bound on the spectral norm of the sample covariance matrix $\|\frac{1}{n}\sum_{i=1}^{n}\vct{x}_i\vct{x}_i^T\|\le 2$ which holds with high probability.

\subsection{Convergence of Quantized SGD (Theorem \ref{theorem2})}
 The proof of this theorem is similar to its counter part in Theorem \ref{theorem1}. We begin by defining $L_{I_{m_k}(t)}(\vct{w})\delequal \frac{1}{m_k}\sum_{j=1}^{m_k}\ell_{i_t^{(j)}}(\vct{w})$. We can repeat the argument of the proof of Theorem \ref{theorem1} and rewrite (\ref{expandmain}) using the updates (\ref{qsgd-updates}). We begin by taking expectations with respect to the randomness in the quanitzation procedure (denoted by $\mathbb{E}_{Q_s}$). This allows us to conclude that
\begin{align*}
    \mathbb{E}_{Q_s}||\vct{w}_t-\vct{w}^*||_2^2=&\mathbb{E}_{Q_s}\bigg[||\vct{w}_{t-1}-\vct{w}^*||_2^2\nonumber-2\frac{\eta}{K}\langle\vct{w}_{t-1}-\vct{w}^*,\sum_{k=1}^{K}Q_s\big(\nabla L_{I_{m_k}(t)}(\vct{w}_{t-1})\big)\rangle\nonumber\\&+\frac{\eta^2}{K^2}||\sum_{k=1}^{K}Q_s\big(\nabla L_{I_{m_k}(t)}(\vct{w}_{t-1})\big)||_2^2\bigg].
\end{align*}
To continue we use a Lemma from \cite{NIPS2017_6768} which shows the stochastic gradient is unbiased and bounds its variance.

\begin{lemma}\label{quant}
 For any vector $\vct{v}\in \mathbb{R}^d$,\\
 (i)\ $\mathbb{E}[Q_s(\vct{v})]=\vct{v}$.\\
 (ii)\ $\mathbb{E}[||Q_s(\vct{v})||^2]\leq\left(1+\text{min}(\frac{d}{s^2},\frac{\sqrt{d}}{s})\right)||\vct{v}||_2^2$.
\end{lemma}

Using Lemma \ref{quant}, we conclude that
\begin{align*}
\mathbb{E}_{Q_s}\big[\langle\vct{w}_{t-1}-\vct{w}^*,\frac{1}{K}\sum_{k=1}^{K}Q_s\big(\nabla L_{I_{m}(t)}(\vct{w}_{t-1})\big)\rangle\big]=&\langle\vct{w}_{t-1}-\vct{w}^*,\frac{1}{K}\sum_{k=1}^{K}\nabla L_{I_{m_k}(t)}(\vct{w}_{t-1})\rangle\\
=&\langle\vct{w}_{t-1}-\vct{w}^*,\nabla L_{I_{m}(t)}(\vct{w}_{t-1})\rangle,
\end{align*}
and
\begin{align*}
\mathbb{E}_{Q_s}\big[||\sum_{k=1}^{K}Q_s\big(\nabla L_{I_{m_k}(t)}(\vct{w}_{t-1})\big)||_2^2\big]=& \mathbb{E}_{Q_s}\bigg(\sum_{k=1}^{K}||Q_s\big(\nabla L_{I_{m_k}(t)}(\vct{w}_{t-1})\big)||_2^2\\&+\sum_{i\neq j}\langle Q_s\big(\nabla L_{I_{m_i}(t)}(\vct{w}_{t-1})\big),Q_s\big(\nabla L_{I_{m_j}(t)}(\vct{w}_{t-1})\big)\rangle\bigg)\\
= & \sum_{k=1}^{K}\mathbb{E}_{Q_s}||Q_s\big(\nabla L_{I_{m_k}(t)}(\vct{w}_{t-1})\big)||_2^2\\&+\sum_{i\neq j} \langle \nabla L_{I_{m_i}(t)}(\vct{w}_{t-1}),\nabla L_{I_{m_j}(t)}(\vct{w}_{t-1})\rangle\\
\leq & \sum_{k=1}^{K}\left(1+\text{min}(\frac{d}{s^2},\frac{\sqrt{d}}{s})\right)||\nabla L_{I_{m_k}(t)}(\vct{w}_{t-1})||_2^2\\&+\sum_{i\neq j} \langle \nabla L_{I_{m_i}(t)}(\vct{w}_{t-1}),\nabla L_{I_{m_j}(t)}(\vct{w}_{t-1})\rangle
\end{align*}
Therefore,
\begin{align*}
    \mathbb{E}_{Q_{s},I_m(t)}\big[||\sum_{k=1}^{K}Q_s\big(\nabla L_{I_{m_k}(t)}(\vct{w}_{t-1})\big)||_2^2\big]\leq & \sum_{k=1}^{K}\left(1+\text{min}(\frac{d}{s^2},\frac{\sqrt{d}}{s})\right)\mathbb{E}||\nabla L_{I_{m_k}(t)}(\vct{w}_{t-1})||_2^2\\&+\sum_{i\neq j} \langle \mathbb{E}\left(\nabla L_{I_{m_i}(t)}(\vct{w}_{t-1})\right),\mathbb{E}\left(\nabla L_{I_{m_j}(t)}(\vct{w}_{t-1})\right)\rangle\\
    \leq & K\left(1+\text{min}(\frac{d}{s^2},\frac{\sqrt{d}}{s})\right)\left(\frac{9dK}{m}+\frac{25}{16}\right)||\vct{w}_t-\vct{w}^*||_2^2\\&+(K^2-K)||\nabla \mathcal{L}(\vct{w}_{t-1})||_2^2\\
    \leq & K\left(1+\text{min}(\frac{d}{s^2},\frac{\sqrt{d}}{s})\right)\left(\frac{9dK}{m}+\frac{25}{16}\right)||\vct{w}_t-\vct{w}^*||_2^2\\&+\frac{25}{16}(K^2-K)||\vct{w}_t-\vct{w}^*||_2^2\\
    \leq & K^2\left(\left(1+\text{min}(\frac{d}{s^2},\frac{\sqrt{d}}{s})\right)(\frac{9d}{m}+\frac{25}{16})+\frac{25}{16}\right)\cdot||\vct{w}_t-\vct{w}^*||_2^2.
\end{align*}
Whence,
\begin{align*}
    \mathbb{E}_{Q_s,I_{m}(t)}||\vct{w}_t-\vct{w}^*||_2^2\leq&||\vct{w}_{t-1}-\vct{w}^*||_2^2-2\eta\langle\vct{w}_{t-1}-\vct{w}^*,\nabla \mathcal{L}(\vct{w}_{t-1})\rangle\nonumber\\&+\eta^2\left(\left(1+\text{min}(\frac{d}{s^2},\frac{\sqrt{d}}{s})\right)(\frac{9d}{m}+\frac{25}{16})+\frac{25}{16}\right)\cdot||\vct{w}_t-\vct{w}^*||_2^2.
\end{align*}
The remainder of the proof is exactly the same as that of Theorem \ref{theorem1}.


%% file: Acknowledgement.tex
\section{Acknowledgements}
M. Soltanolkotabi is supported by the Packard Fellowship in Science and Engineering, a Sloan Research Fellowship in Mathematics, an NSF-CAREER under award \#1846369, the Air Force Office of Scientific Research Young Investigator Program (AFOSR-YIP)
under award \#FA9550-18-1-0078, an NSF-CIF award \#1813877, and a Google faculty research award.

%% file: Appendix.tex
\section{Appendix}\label{appen}

In the statement of the two Theorems, the probability of success depends on the proximity of the initial estimate to the solution.  To alleviate this dependency we directly adapt the \textit{Ensemble} Algorithm used in \cite{doi:10.1093/imaiai/iay005}. See Algorithm \ref{Alg1} for details of this approach.

We now state a result regarding the performance of the ensemble method. We note that the proof of this lemma is essentially identical to its counterpart in \cite{doi:10.1093/imaiai/iay005} and requires only minor modifications.

\begin{proposition}{(Guarantees for ensemble methods).} Consider the setup and assumptions of Theorem \ref{theorem1}. Furthermore assume that $\text{Prob}(\text{failure})\leq 1/3$. Then, for any $\delta'>0$ there is an absolute constant $C$ such that if $L\geq C\log{(1/\delta')}$ then the estimate $\vct{\hat{w}}$ obtained via Algorithm \ref{algorithm1} satisfies $||\vct{\hat{w}}-\vct{w}^*||_2\leq 9\epsilon||\vct{w}_0-\vct{w}^*||_2$.

\end{proposition}


\begin{algorithm}[H]
    \caption{Ensemble Method}\label{algorithm1}
    \hspace*{\algorithmicindent} \textbf{Input:} Feature vectors $\vct{x}_1,\vct{x}_2,...,\vct{x}_n,$ labels $y_1,y_2,...,y_n$, relative error tolerance $\epsilon$, iteration count $K$ to achieve the desired error $\epsilon$, trial count $L$.\\
    \hspace*{\algorithmicindent} \textbf{Output:} An estimate $\vct{\hat{w}}$ for $\vct{w}^*.$ 
    \begin{algorithmic}[1]
    
   
    \State  Initialize $\vct{w}_0$ to satisfy the condition of Theorems \ref{theorem1} and \ref{theorem2}.
    \State \textbf{for} $l=1,...,L$ run $K$ SGD updates from the initial point $\vct{w}_0$ to obtain $\vct{w}_K^{(l)}$.
    \For{$l=1,...,L,$}
    \If{$|B(\vct{w}_k^{(l)},2\sqrt{\epsilon})\cap \{\vct{w}_K^{(1),...,\vct{w}_K^{(L)}}\}|\geq L/2$} 
    \State \textbf{Return} $\vct{\hat{w}}:=\vct{w}_K^{(l)}$
    \EndIf
    \EndFor
    \end{algorithmic}
    \label{Alg1}
    \end{algorithm}